# AN INTEGER PROGRAMMING MODEL FOR THE DYNAMIC LOCATION AND RELOCATION OF EMERGENCY VEHICLES: A CASE STUDY

**Mahdi Moeini, Zied Jemai, Evren Sahin**
Laboratoire Génie Industriel (LGI), Ecole Centrale Paris,
Grande Voie des Vignes, F-92 295, Châtenay-Malabry, France.
{mahdi.moeini, zied.jemai, evren.sahin}@ecp.fr

**Abstract:** In this paper, we address the dynamic Emergency Medical Service (EMS) systems. A dynamic location model is presented that tries to locate and relocate the ambulances. The proposed model controls the movements and locations of ambulances in order to provide a better coverage of the demand points under different fluctuation patterns that may happen during a given period of time. Some numerical experiments have been carried out by using some real-world data sets that have been collected through the French EMS system.

## 1  INTRODUCTION

Due to the crucial role of the Emergency Medical Services (EMS) in saving lives, numerous studies have been carried out in order to improve the quality of the EMS systems. In this context, different research directions have been taken into account. Some examples are adaptation of the service modes to the changes in the customer needs (such as home care services), personnel scheduling in the medical centres, location of the service centres, etc. In any case, two main objectives are saving the lives (by reducing the mortality in the emergency cases) and reducing the costs.

Among the EMS literature, the problem of locating the emergency service vehicles has attracted special attention during decades of research. The vehicle location problem in the context of EMS systems is dealing with locating the vehicles in some potential service sites in order to reduce the delay of covering the emergency service demands.

Each emergency service vehicle is completely equipped to all emergency facilities that medical team may need in their missions. Due to this fact, it is quite expensive to buy any of these emergency vehicles. Consequently, any emergency service has access to a limited number of emergency vehicles; hence, it is important to optimally locate them in order to improve the responsiveness of the system.

### 1.1   Literature review

The earliest EMS models have been introduced in 70s by Toregas et al. [7]. During decades the location problems of EMS vehicles became an active research area and numerous papers have been published on this topic. The published papers may be classified according to their nature: static, dynamic, or stochastic models.

Toregas et al. [7] introduced the Location Set Covering (LSC) model that minimizes the number of the necessary ambulances for covering all demand points. The LSC model can penalize the users of the model by its expensive solutions; because it may provide a necessary number of ambulances that is too larger than it would be. Furthermore, the LSC model is so rather basic and it does not permit location of more than one ambulance in a service centre.

Due to the limits of the LSC model, the Maximal Covering Location problem (MCLP) has been presented by Church et al. [3]. The MCLP model tries to maximize the covered population by taking into account a predefined number of ambulances.

The LSC and MCLP models are static models, in the sense that they do not take into account the possible fluctuations in the EMS system. In fact, when a call arrives to the call centre of the EMS service centre, it may need affectation of an ambulance. If such need is confirmed, an ambulance will be affected to cover the demand point. At this stage, the corresponding ambulance will be no longer available. Consequently, the static models must be solved from scratch for a smaller number of ambulances. This procedure is computationally expensive. At this point, one may use the dynamic models.

Another inconvenience of the LSC and MCLP models is due to the problem of simultaneous emergency calls. More precisely, it is possible to receive more than one emergency service demand call at the same time or in a very short time delay. Each of the service demand points must be covered; consequently, we may need to support the zones by more than one ambulance. The classical LSC and MCLP models are not able to provide such service.

In order to overcome the inconveniences of the static models, several approaches have been introduced in the literature. One approach consists of employing more than one ambulance to cover the simultaneous emergency demand calls.

The double standard model (DSM) [4] is an example of the models that use multiple ambulances in covering the demands. The DSM model is based on the assumption that all demands must be covered by an ambulance within $r_2$ minutes and a proportion $\alpha$ of the total demand must be covered within $r_1$ minutes. Obviously, $r_2 > r_1$.

The multi-coverage models try to handle the problem of the uncertainty in the demands. Stochastic programming is another approach that is used to take into account the uncertainty. In spite of the multi-coverage models, the stochastic programming models try to cover the uncertainty in a more explicit way. In the stochastic models, the origins of uncertainty are considered to be either the availability of the ambulances (vehicles) or the occurrence of the service requests at the demands points (see [1], [2], [6]).

Finally, due to the dynamic setting of the EMS systems, dynamic optimization models seem to be suitable choices in efficient covering of the EMS demand points. In this esprit, Gendreau et al. [5] has introduced a dynamic redeployment (relocation) problem ($RP^t$). This model is as an extension to the DSM [4]. The model ($RP^t$) maximizes the number of the demand points that are covered two times and minimizes the costs associated to the movements of the vehicles. The model contains a penalty parameter that takes all dynamic changes into account.

## 1.2 Results

In this paper, we are interested in proposing a new model that fits to the French EMS system. The new model is based on the $RP^t$ (see [5]) and we show that (in the context of the French EMS system) the proposed model is more efficient than $RP^t$. In order to formalize the model, we introduce a new parameter into the $RP^t$ model. We believe that the new parameter improves the ability of the model in covering the emergency demands. The parameter is computed by using different fluctuation patterns of emergency demands during a given period of time.

Once the model is built, we need to verify its abilities in covering the emergency demands. Hence, the presented model is compared to the $RP^t$ in terms of the ability in covering the emergency service requests. The comparisons are based on the experiments that have been carried out by using some real-world data sets. They have been collected through the French emergency medical service system. According to the numerical results, the proposed model provides a better coverage of the emergency demands.

The structure of the paper is as follows. The $RP^t$ model [5] is reviewed and our dynamic model is presented in Section 2. The models are tested on real-world data sets that have been collected through the French EMS system. The computational experiments are reported in Section 3. Finally, the last section includes some conclusions.

## 2 DYNAMIC LOCATION AND RELOCATION MODEL

In this section, we present our dynamic location and relocation model. The proposed model can be considered as an extended version of the classical $RP^t$ model that has been introduced by Gendreau et al. [5].

In the EMS' context, each service point covers some demand points (zones). One or more vehicles are associated to each service point and they are responsible to cover the demands. In some circumstances, one may need to cover a demand point (zone) by more than one vehicle. This is due to the curse of uncertainty and it is related to the potential of a point (zone) in producing more than one emergency requests during a specific period of time. This situation corresponds to reception of an emergency service request while the covering vehicle is busy because of giving service to another demand in the *same zone*. We will call these demands as *simultaneous* emergency service demands versus the *simple* emergency service requests (that correspond to the demands arriving during the availability of the vehicle). There will be a conflict if the two demand points (with emergency needs) are located in the same zone and are supposed to be covered by the same service point. We address this situation by introducing some parameters into the model. These parameters are computed by using the historical emergency demands' data of each zone.

### 2.1 The Dynamic Relocation Problem $(DRP^t)$

For the sake of completeness, we start by describing the classical $RP^t$ model of Gendreau et al. [5]. In order to present the model, we will use the notations that are summarized in Table 1.

*Table 1: Notations: parameters and decision variables.*

| --- | *Description* |
|---|---|
| $i \in I := \{1,...,n\}$ | $i$ is a demand point and $I$ is the set of all potential demand points. |
| $j \in J := \{1,...,m\}$ | $j$ is a service point (centre) and $J$ denotes the set of all service centres. |
| $k \in K := \{1,...,|K|\}$ | $k$ is an ambulance and $K$ is the set of all ambulances. |
| $U_j$ | the maximum number of the ambulances that can be assigned to the service centre $j$. |
| $d_i$ | denotes the mean density of the emergency demands at the point $i$. |
| $d_i^1, d_i^2$ | mean density of the emergency service demands at the point $i$ for a simple ($d_i^1$) or simultaneous ($d_i^2$) call. |
| $r_1, r_2$ | the time thresholds to be respected in covering any demand point ($r_1 < r_2$). |
| $\gamma_{ij}$ | a binary parameter that denotes whether a demand point $i$ is accessible from the service centre $j$ in $r_1$ minutes. |

| $\delta_{ij}$ | a binary parameter that denotes whether a demand point $i$ is accessible from the service centre $j$ in $r_2$ minutes. |
|---|---|
| $\alpha \in [0,1]$ | a real number indicating the proportion of all emergency service demands that must be covered in a given delay. |
| $M^t_{jk}$ | a real valued parameter for controlling the relocations and movements of the vehicles at each period $t$; particularly, $M^t_{jk}$ takes larger values in order to prevent long-distance travels of the vehicles. |
| $x^\lambda_i \in \{0,1\}$ | to say whether the demand point $i$ is covered $\lambda$ times (for $\lambda \in \{1,2\}$). |
| $y_{jk} \in \{0,1\}$ | to say whether the ambulance $k$ is located in the service point $j$. |

Associated to a given time $t$, there is a real valued penalty parameter $M^t_{jk}$ that is incorporated into the model. This parameter plays an important role in the dynamic structure of the model and in the stability of the provided location plans throughout the day (or the operational period of the model). In fact, for a given $t$, the penalty parameter $M^t_{jk}$ is associated to the relocation of ambulance $k$ from its current position to service point $j \in J$. The value of this parameter is adjusted at any time $t$ according to the different information regarding the vehicle $k$ and the service point $j$. This information may contain frequent moves of the vehicle in the past, round trips, and relocations over long distances.

By using the presented notations, the model $RP^t$ of Gendreau et al. [5] reads as follows:

$$\max \sum_{i=1}^{n} d_i x_i^2 - \sum_{j=1}^{m} \sum_{k=1}^{|K|} M^t_{jk} y_{jk} \qquad (1)$$

$$\text{Subject to: } \sum_{j=1}^{m} \sum_{k=1}^{|K|} \delta_{ij} y_{jk} \geq 1 : \forall i \qquad (2)$$

$$\sum_{i=1}^{n} d_i x_i^1 \geq \alpha \sum_{i=1}^{n} d_i \qquad (3)$$

$$\sum_{j=1}^{m} \sum_{k=1}^{|K|} \gamma_{ij} y_{jk} \geq x_i^1 + x_i^2 : \forall i \qquad (4)$$

$$x_i^1 \geq x_i^2 : \forall i \qquad (5)$$

$$\sum_{j=1}^{m} y_{jk} = 1 : \forall k \qquad (6)$$

$$\sum_{k=1}^{|K|} y_{jk} \leq U_j : \forall j \qquad (7)$$

$$x_i^1, x_i^2 \in \{0,1\} : \forall i \text{ and } y_{jk} \in \{0,1\} : \forall j, \forall k. \qquad (8)$$

In this model, the objective is to maximize the demand points that are covered two times and to minimize the costs related to the relocation of the vehicles. The constraint (2) ensures the absolute coverage of the demands within $r_2$ units (of time per minutes). The requirements related to the partial covering of the demands are expressed by the constraints (3) and (4). According to the constraint (3), $\alpha\%$ of all emergency demands is covered. The constraint (4) states that the number of vehicles waiting in $r_1$ units (of time per minutes) from the demand point $i$ must be either at least equal to 1, if $x_i^1$ is equal to 1, or at least 2 if $x_i^1 = x_i^2 = 1$. Constraints (5) say that any demand point $i$ can be covered twice if and only if it is already covered at least once. According to the constraint (6), each ambulance must be assigned to a service centre. Finally, an upper bound is defined, by the constraints (7), on the number of the ambulances that can be assigned to a service point. Constraints (8) are the integrality constraints.

The model $(RP^t)$ has been successfully applied in different countries. In spite of this fact, it can be changed in order to be casted into our case study in the context of the French EMS system. In fact, our case study has been carried out on Val-de-Marne (that is a county in France) where the intensity of emergency service demands is not high. In spite of this fact and in order to take into account the uncertainties, we need to provide the best possible coverage of the emergency demand points. To this aim, we introduce some new parameters in order to take into account different kinds of coverage. According to a given demand point, one may need to give more importance to the double coverage in comparison to another point. This fact is included explicitly in the new model. In order to present our model, we need to define two new parameters. We will use the parameters $d_i^1$ and $d_i^2$ for specifying, respectively, the mean occurrence number (intensity) of the all service requests and the mean occurrence number (intensity) of the simultaneous emergency demands at the demand point $i$. We summarize our Dynamic Relocation Problem $(DRP^t)$ as follows:

$$\max \sum_{i=1}^{n} \left( d_i^1 x_i^1 + d_i^2 x_i^2 \right) - \sum_{j=1}^{m} \sum_{k=1}^{|K|} M_{jk}^t y_{jk} \tag{10}$$

Subject to: $\quad \sum_{i=1}^{n} d_i^1 x_i^1 \geq \alpha \sum_{i=1}^{n} d_i^1 \tag{12}$

and the constraints (2), (4) – (8).

In this model, the objective is to maximize the demand points that are covered and to minimize the costs related to the relocation of the vehicles. In order to cover the demand points, the model takes into account the weights associated to the two categories of service demands: $d_i^1$ and $d_i^2$ (see Table 1 for more precise definitions of $d_i^1$ and $d_i^2$). According to the constraint (12), $\alpha\%$ of all emergency demands is covered. The remaining constraints of the model are the same as the model $(RP^t)$.

The differences between the models $(RP^t)$ and $(DRP^t)$ are essentially in the objective functions and also in the proportional coverage constraints. In a similar way to the $(RP^t)$, the $(DRP^t)$ maximizes the coverage of the emergency demands but the $(DRP^t)$ model uses two parameters $d_i^1$ and $d_i^2$. Due to the randomness of the demands, there may happen some situations during which some simultaneous demands occur. By enforcing double coverage of the demand points, we can cover this kind of situations. To this aim, the parameter $d_i^2$ is used.

## 3 COMPUTATIONAL EXPERIMENTS

### 3.1 Data description

The models were used for the EMS system in Val-de-Marne, a county in France. The population of the county amounts to approximately 1.3 million inhabitants and it is divided into 47 quarters.

The EMS call centre of the county of Val-de-Marne receives more than 1000 calls per day, but just a small part of the calls require sending an EMS vehicle; that is, in general, between 20 and 30 calls per day. In our experiments, we suppose that each of the calls must be covered in less than 10 minutes and 8 ambulances are in use in the county. Furthermore, in our experiments, we suppose that 12 centres are in daily use (see Figure 1).

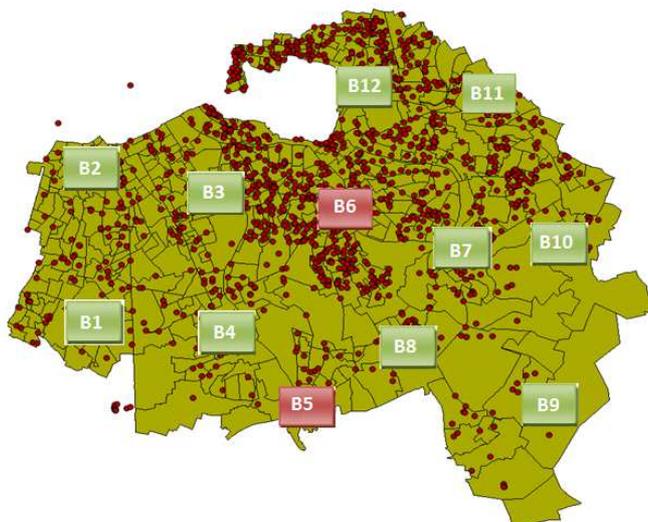

*Figure 1: The 12 service centres (B1-B12) in the county of Val-de-Marne. B5 and B6 are currently operational and there is a plan to use the other centres.*

For our computational experiments we used some recently collected data from the EMS system of the county. Data collection has been made possible by means of a new GPS localization system. It has been installed in the Hospital *Henri Mondor* that is located in the county of Val-de-Marne.

The standard solver IBM Cplex (version 12.2) has been used to solve the mathematical optimization models corresponding to the case study. Since the size of the models is not large, the models are solved in a quite short time, which is less than 2 seconds.

In order to compare the models $(RP^t)$ and $(DRP^t)$, we solved them under same conditions by using the same data sets. The performance of the models is measured by means of their ability in covering the EMS service demand points.

Different proportional covering percentages (i.e., $\alpha$) have been taken into account. In fact, in our experiments, $\alpha$ varies from 90 % to 100 %.

### 3.2 Results

Figures 2 and 3 show the experiments that have been carried out on two consecutive time periods. The figures show the coverage proposed by the models $(RP^t)$ and $(DRP^t)$ (shown on the figures by $RP$ and $DRP$, respectively).

Figure 2 shows the results for the starting time period. At this period the values of $M^t_{jk}$ are all initialized by zero (for all $j,k$).

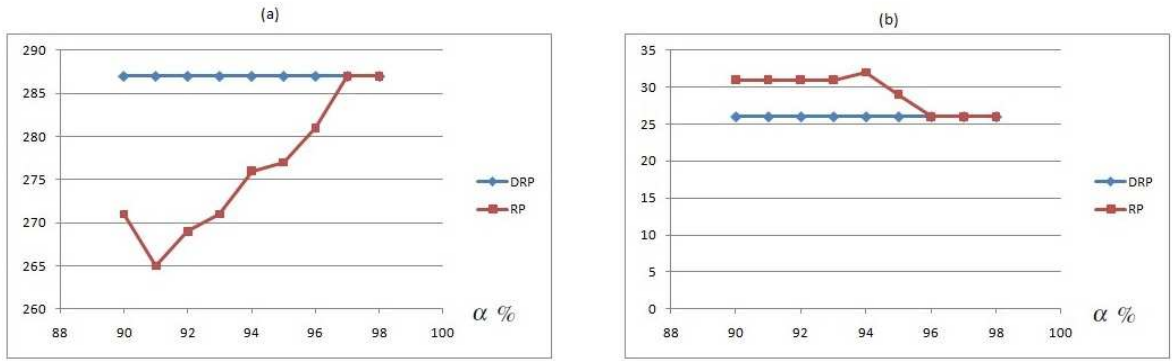

*Figure 2: The number of the single (a) and double (b) covered demands (vertical axis) for different proportions of the total demands (horizontal axis).*

**Some observations on the results of the first period:**

- For a given proportion of demand to be satisfied, the behaviour of the two models is significantly different. The value of the covered demands remains stable in the $(DRP^t)$ model, but this value may decrease or increase in the $(RP^t)$ model. This observation can be justified by reviewing the structure of the objective functions. In fact, the variables $x_i^1$ and $x_i^2$ (weighted by $d_i^1$ and $d_i^2$, respectively) are present in the objective function of $(DRP^t)$, but this is not the case of $(RP^t)$.
- According to Figure 2, the $(DRP^t)$ presents an ambulance deployment policy with a better coverage in comparison to the $(RP^t)$. The coverage includes all types of the calls, i.e., simultaneous demands as well as the non-simultaneous (i.e., simple) ones. In contrary to Fig. 2 (a), Fig. 2 (b) shows a better coverage provided by the $(RP^t)$ model. In a similar way to the previous case, the difference is justified by the structure of the objective functions in the $(DRP^t)$ and the $(RP^t)$. The $(RP^t)$ model includes only the $x_i^2$ (that is weighted by $d_i^1$), which privileges the double coverage.

In order to pass from the first period to the second period, we need to adjust the values of $M_{jk}^t$ (where $j$ indicates a service centre and $k$ is a given EMS vehicle). The main issue is to reduce the movements of the vehicles. Based on this policy, the distance between service centres has been considered as the value of $M_{jk}^t$. Furthermore, we suppose that in the second period one of the vehicles is busy because of a mission. Hence, we must solve the optimization models with one vehicle less than the previous period, i.e., $k = 7$. The results are depicted in Figure 3.

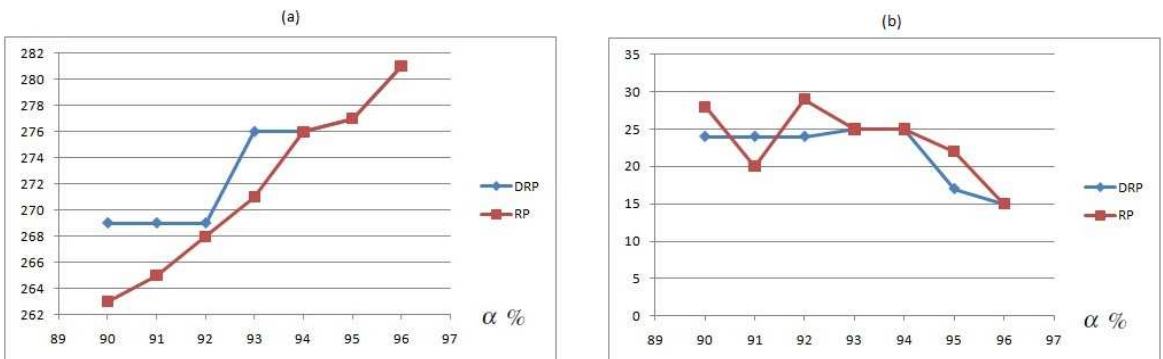

*Figure 3: The number of the single (a) and double (b) covered demands for the Period 2. The vertical axis corresponds to the covered demand points and the horizontal axis shows the different values for $\alpha$.*

**Some observations on the results of the second period:**

- The results of the second period are significantly different from the results of the first period. We remember that the values of $M_{jk}^{t}$ are adjusted in a way to reduce the total movements of the ambulances. Due to this fact, when we consider two $(DRP^{t})$ models corresponding to two different values of $\alpha$, the corresponding objective functions of the models may be different. Indeed, *any similarity in the solutions of the first period may provide similar models for the second period.*
- Similarly to the first period, we observe that the $(DRP^{t})$ model provides solutions for which we have a better coverage of the *simple* demands. Furthermore, there is no more absolute superiority in the quantity of the *double* covered demands by the $(RP^{t})$ model in comparison to the $(DRP^{t})$ model.

## 4 CONCLUSIONS

In this paper, we presented a new dynamic location and relocation model in the context of the Emergency Medical Services. The model has been tested and verified on real-world data sets. According to the experiences, the model is solved efficiently for the studied cases. In spite of this fact, one will need some more efficient approaches for solving the large-scale programs. A set of experiments has been carried out to emphasise usefulness of the proposed model. To this aim, the model has been compared to one of the classical existing models. The numerical results show improvements in the coverage of the demands by using the introduced model.

**Acknowledgments**
This work has been supported by the French National Agency of Research (ANR) under the contract *Performance Optimization of SAMU* (ANR - POSAMU).

**References**
[1] Ball M.O., Lin L.F., 1993. A reliability model applied to emergency service vehicle location, Operations Research, Vol. 41, pp. 18-36.
[2] Beraldi P., Bruni M.E., Conforti D., 2004. Designing robust emergency medical service via stochastic programming, European journal of Operational Research, 158, 183-193.
[3] Church R. L., ReVelle C.S., 1974. The maximal covering location problem, Papers of Regional Science Association, Vol. 32, pp. 101-118.
[4] Gendreau M., Laporte G., Semet F., 1997. Solving an ambulance location model by tabu search, Location Science, Vol. 5, pp. 75-88.
[5] Gendreau M., Laporte G., Semet F., 2001. A dynamic model and parallel tabu search heuristic for real-time ambulance relocation, Parallel Computing, Vol. 27, pp. 1641-1653.
[6] ReVelle C. S. et K. Hogan., 1989. The maximum availability location problem, Transportation Science, Vol. 23, pp. 192-200.
[7] Toregas C., Swain R., ReVelle C.S., Bergman L., 1971. The location of emergency service facilities, Operations Research.